\documentclass{article}

\usepackage[preprint]{neurips_2024}
\usepackage[utf8]{inputenc} 
\usepackage[T1]{fontenc}    
\usepackage{hyperref}       
\usepackage{url}            
\usepackage{booktabs}       
\usepackage{amsfonts}       
\usepackage{nicefrac}       
\usepackage{microtype}      
\usepackage{xcolor}         
\usepackage{caption}
\usepackage{subcaption}
\usepackage{graphicx}
\usepackage{times}
\usepackage[inline]{enumitem}

\usepackage{amsfonts}       
\usepackage{nicefrac}       
\usepackage{microtype}      
\usepackage{xcolor}         
\usepackage{booktabs}
\usepackage{amsmath}
\usepackage{multirow}
\usepackage{enumitem}
\usepackage{rotating}
\usepackage[normalem]{ulem}
\usepackage{amssymb}
\usepackage{colortbl}
\usepackage{linguex}
\usepackage{wrapfig}

\usepackage{siunitx} 
\usepackage{multirow, makecell}
\usepackage{pifont}
\usepackage{tabularx}
\usepackage[all]{nowidow}
\definecolor{darkgreen}{rgb}{0.0, 0.42, 0.24}
\definecolor{green}{RGB}{112, 173,71}
\definecolor{blue}{RGB}{68, 114,196}
\definecolor{orange}{RGB}{237, 125,49}
\definecolor{red}{RGB}{202, 54,49}
\definecolor{yellow}{RGB}{222,194, 142}
\usepackage[linesnumbered,ruled,vlined]{algorithm2e}
\SetKwInput{KwInput}{Input}  
\usepackage{tikz}
\newcommand{\cmss}[1]{{\fontfamily{cmss}\selectfont{#1}}}
\newcommand{\method}{\cmss{ProxyNN}\xspace}

\newcommand{\ignore}[1]{}

\usepackage{xspace}
\usepackage{hyperref}

\title{Optimization Landscapes Learned: Proxy Networks Boost Convergence in Physics-based Inverse Problems}

\author{%
  Girnar~Goyal$^{1}$, Philipp~Holl$^{2}$, Sweta~Agrawal$^{3}$, Nils~Thuerey$^{2}$  \\
  $^{1}$Ludwig Maximilian University of Munich, \\
  $^{2}$Technical University of Munich, $^3$Instituto de Telecomunicações \\
  \texttt{girigoyal@gmail.com} \\
}

\begin{document}
\maketitle

\begin{abstract}
 Solving inverse problems in physics is central to understanding complex systems and advancing technologies in various fields. Iterative optimization algorithms, commonly used to solve these problems, often encounter local minima, chaos, or regions with zero gradients. This is due to their overreliance on local information and highly chaotic inverse loss landscapes governed by underlying partial differential equations (PDEs). In this work, we show that deep neural networks successfully replicate such complex loss landscapes through spatio-temporal trajectory inputs. They also offer the potential to control the underlying complexity of these chaotic loss landscapes during training through various regularization methods. We show that optimizing on network-smoothened loss landscapes leads to improved convergence in predicting optimum inverse parameters over conventional momentum-based optimizers such as \textsc{BFGS} on multiple challenging problems.
\end{abstract}

\section{Introduction}
\label{introduction}

Solving inverse problems is pivotal across scientific and engineering domains, with applications ranging from optimizing fluid dynamics \citep{OULGHELOU2022105490} and materials design \citep{Fung2021} to enhancing structural health monitoring \citep{8103129}, manufacturing optimization \citep{WURTH2023112034}, and weather prediction \citep{Sixun_Huang_2005}. Often, they are difficult to solve as they involve determining the causes of observed effects, and even minor perturbations in the output values can cause significant variations in the reconstructed input. This is why solving inverse problems is a substantially more challenging task than observing the effects themselves in a forward problem.



Various techniques, such as Bayesian inference \citep{shahriari2015taking}, quasi-Newton \citep{ruder2017overview, broyden1970convergence}, and machine learning methods \citep{2021arXiv210515044C, 2022NJPh...24f3002C,mohammad2021regularization, Antil_2023}, have been devised to solve the inverse problems. Popular iterative methods based on approximations of the Hessian matrix, such as the \textsc{BFGS} \citep{broyden1970convergence}, often encounter convergence issues in local optima, flat, and chaotic regions of the underlying non-linear inverse loss landscapes. While the local optima attract the optimizers, flat and chaotic regions frequently create directional traps which makes these loss landscapes extremely difficult to optimize. Step size adjustments and the addition of momentum do not fully overcome these challenges.



Recent research investigates using deep neural networks (DNN) as surrogates for the inverse problem governed by non-linear partial differential equations (PDEs), aiming to recover the simulation inputs through end-to-end predictions \citep{PFROMMER2018426, ren2021benchmarking}.  However, using DNNs for complex inverse problems is challenging due to: (i) limited and noisy data creating the risk of overfitting during training (ii) a lack of understanding of the network learning mechanism \citep{make3040048}, and (iii) a lack of generalization across inverse problems \citep{Goodfellow-et-al-2016}.

We instead hypothesize that, when provided with a true spatio-temporal trajectory for which the inverse solutions are sought, a \textit{proxy neural network} (\method) can be directly trained to predict the \textit{\textbf{configuration loss}} that measures the deviation of this trajectory from that induced by randomly sampled target parameters from the same initial state.
However, the non-linearity of these configuration loss landscapes would still remain the core issue in finding optimum solutions to the inverse problems. With our work, we aim to answer the following research questions:

\textit{Can we use deep neural networks to model and predict the target configuration loss landscapes? How can we utilize the training process of \cmss{ProxyNNs} to simplify the complexity of the target configuration loss landscapes?}

Motivated by the work of \citet{dherin2022neural} who utilize regularization methods for controlling the network's complexity during training, we aim to gain control over the underlying complexity of the configuration loss landscapes via several regularization techniques.

%
We show that \cmss{ProxyNNs} can successfully predict chaotic, non-linear loss landscapes in inverse problems and present the opportunity to steer the underlying complexity during training through various regularization techniques. Under regularization pressure, \cmss{ProxyNNs} learn to predict simpler solutions to such complex landscapes while preserving the fundamental characteristics of the underlying PDEs. We find that optimizing configuration loss landscapes predicted by complexity-controlled \cmss{ProxyNNs} yield better convergence through momentum-based optimizers such as \textsc{BFGS}. We show the improvement in accuracy of the proposed \method-based optimization over traditional methods in three complex inverse problems: (i) the Inviscid Burgers\' equation, representing fluid flow dynamics; (ii) a chaotic system following the Kuramoto-Sivashinsky equation; and (iii) a rigid body N-dimensional simulation inspired by Billiards. 


\section{Method}

Several studies have emphasized the potential of DNNs as effective surrogates for numerical solvers \citep{WURTH2023112034, 10.1145/2939672.2939738, 2020arXiv200408826D, 10.1007/978-3-030-58309-5_20, PFROMMER2018426, DBLP:journals/corr/abs-1806-02957, MICHOSKI2020193, ANANTHAPADMANABHA2021110194, SHEN2022110460}. In this work, we demonstrate that the surrogate capabilities of neural networks can be further expanded to approximate intricate relationships between the system trajectory and predicted trajectories evolved through randomly sampled control parameters.
We begin by formulating a generalized inverse problem framework where we have access to the spatio-temporal evolution of a physical system, and we aim to recover the influencing control parameters.

\subsection{Generalized Landscape for Inverse Problems}

In the forward problem, the evolution of an initial physical state $Y_0$ under the influence of control parameters $X^*$ is obtained by numerically solving the underlying partial differential equation $\mathcal{P}$ as:
\begin{equation*}
Y^* = \mathcal{P}(Y_0, X^*)
\end{equation*}

In an inverse problem, our goal is to determine the unknown control parameters $X^*$ from the known trajectory of states $Y^*$. The key is to quantify how these control parameters influence the spatio-temporal evolution of physical states through underlying physical laws and constraints. This can be done by comparing the trajectory evolved under random values of control parameters $X_s \in Z$, from the initial state $Y_0$: $Y_s = \mathcal{P}(Y_0, X_s)$.
Both the \textit{true trajectory} $(Y^*)$ and the \textit{comparison trajectory} $(Y_s)$ are numerically 
computed with 
the same initial state $Y_0$ but different control parameters $X^*$ and $X_s$ respectively. To quantify the influence of the control parameters, we define the \textit{Configuration Loss} function as:
\begin{equation}
   \mathcal{L}(Y^*, X_s) = ||\mathcal{P}(Y_0, X_s) - Y^*||_2^2
   \label{eq:config_loss}
\end{equation}

Optimization of the \textit{Configuration Loss} function in the control parameter space (${X_s \in Z}$) would theoretically yield the solution to the inverse problem:
\begin{equation} 
    X^* = min_{X_s \in Z}(\mathcal{L}(Y^*, X_s))
    \label{eq:predicted_true_config_loss}
\end{equation}

\begin{wrapfigure}{o}{0.54\textwidth}
    \centering
    \includegraphics[width=0.54\textwidth]{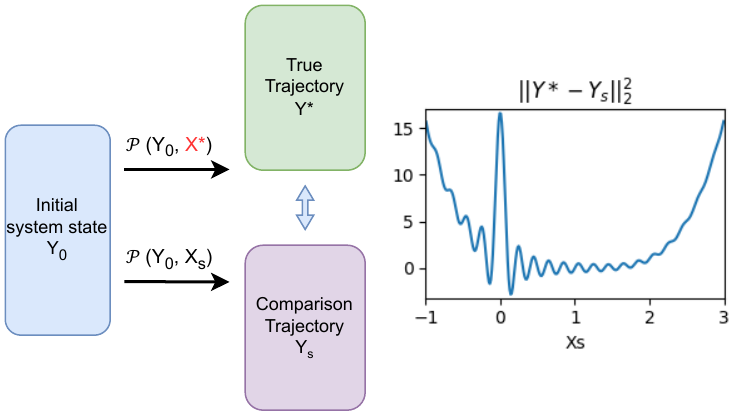}
    \caption{Schematic representation of formulating the \textit{configuration loss} landscape $\mathcal{L} = ||Y_s - Y^*||^2$. The Gramacy and Lee function represent a test case for configuration loss landscapes in this figure.}
\label{fig:scheme_inverse_landscape}
\end{wrapfigure}

In this framework, we expect the configuration loss landscape to obtain a global minimum at the true parameter set value, i.e. $\mathcal{L}(Y^*, X_s)|_{X_s = X^*} = 0$. Depending on the complexity of the physical system and underlying \textsc{PDE}s, the landscape can be geometrically chaotic with multiple local minima and regions of sharp and vanishing gradients. These issues encountered during the optimization of such loss landscapes can be also understood through a popular test case, the Gramacy \& Lee function, $f(x) = \frac{\sin(10\pi x)}{2x} + (x - 1)^4$ \citep{gramacy2010optimization}. It contains multiple local minima along with a global minimum at $(X_{s} = 0.143)$ in $[-1,3]$ (See Figure~\ref{fig:scheme_inverse_landscape}). The local-minima problem causes poor convergence in iterative momentum-based optimizers such as BFGS. Depending on the initial guess of the control parameters, BFGS often yields a piece-wise convergence and fails to converge to the optimum parameters.

In this work, we propose to first replicate the configuration loss landscape, $\mathcal{L}(Y^*, X_s)$, via DNNs which are 
established as universal function approximators 
(Section~\ref{subsec:losssurrogate}). However, this would still require finding the global minimum using the model-predicted loss. To sidestep this, we instead aim to gain control over the complexity of the inverse configuration loss landscape during the training of a proxy network via implicit and explicit regularization techniques (Section~\ref{complexity_control}) motivated by the work in \citet{dherin2022neural}.

\subsection{\cmss{ProxyNNs} for the ``Configuration Loss'' Function} \label{subsec:losssurrogate}

The complex geometric features in the \textit{Configuration Loss} landscape motivate the exploration of using machine learning methods. We employ Convolutional Neural Networks (CNN) \citep{WINOVICH2019263} to train deep proxy neural networks, anticipating that the network can effectively grasp the inherent partial differential equation (PDE) and accurately emulate the configuration loss, $(\mathcal{L})$.

For a physical system with a trajectory of states, $Y^*$, governed by PDE, $\mathcal{P}$, \method is trained to map unrelated inputs, $Y^*$ and randomly sampled $X_s \in Z$, to the target configuration loss $\mathcal{L}$. The network is then trained by minimizing the network training L2 loss, $\mathcal{L}_N$ using the Adam optimizer \citep{kingma2017adam} in each step as:
\begin{equation}
   \mathcal{L}_N = ||f_\theta(Y^*, X_s) - \mathcal{L}(Y^*, X_s)||^2
   \label{eq:training_loss}
\end{equation}

where $\mathcal{L}(Y^*, X_s)$ and $f_\theta(Y^*, X_s)$ represent the ground truth (equation~\ref{eq:config_loss}) and proxy network predicted configuration loss values for the true trajectory $Y^*$.

\subsection{Modulating \method's Complexity through Regularization} 
\label{complexity_control}

This section outlines our approach for controlling the intricacies of learning the landscape of inverse configuration loss. We achieve this by employing a range of implicit and explicit regularization techniques throughout the training of the \method.

\textbf{Noise regularization}: The networks are trained to replicate the configuration loss function through noise-labeled inputs  ($X_s + \sigma \mathcal{N})$ where $\sigma$ is the scaling hyperparameter for the noise sampled from a Gaussian distribution $\mathcal{N}$ \citep{blanc2020implicit, dherin2022neural}. The introduction of scaled Gaussian noise puts regularization pressure on the network learning process, which enhances the generalizing capacity of networks and helps mitigate the impact of high-frequency signals within the target configuration loss function.


\textbf{Loss penalty-based regularization}: We observe that the high-frequency geometric features and regions containing maxima or minima are identified at later training stages of the network. Hence, we introduce a training-loss penalty that creates a geometric-bias-based regularization during the network training, favoring low-lying regions of the target configuration loss landscape. The network training loss (equation~\ref{eq:training_loss}) is scaled with the loss-penalty hyperparameter $\mu > 1: \forall \, (X_s \in Z) \text{ where}  \, f_\theta(Y^*, X_s) > \mathcal{L}$. This imposes a regularization influence to encourage improved learning of geometric features near regions that contain minima.


The regularization techniques with hyperparameters $\{ \sigma, \mu\}$ can be used to achieve reasonable control over the \method's complexity with the modified network training loss as:
\begin{equation}
\begin{split}
    \mathcal{L}^{R}_N =  \mu \times || f_\theta(Y^*, X_s + \sigma\mathcal{N}) - \mathcal{L} (Y^*, Y_s) ||_2^2 
\end{split}
\label{eq:regularized_training_loss}
\end{equation}

The full algorithm to train \cmss{ProxyNNs} is shown below:
\begin{algorithm}

\KwInput{Ground truth forward simulated trajectories with their true inverse values, $D= \{Y^*, X^*\}_{i=1}^N$, Numerical simulation framework for the underlying PDE, $\mathcal{P}$ }

   \For{training step $t=1...T$}
   {
        Sample state trajectory $Y^*$ from dataset $D$ and extract its initial state $Y_0$\\
         \For{i in $1 .. n$}
           {
            Randomly sample control parameters, $X_s \in Z$\\
            From $Y_0$, simulate comparison trajectory $Y_s$ evolved under $X_s$, \textit{i.e.}, $Y_s = \mathcal{P}(Y_0, X_s)$ \\
            Compute the true configuration loss between the \textit{true trajectory} and \textit{comparison trajectory}, $\mathcal{L} = ||Y^* - Y_s||^2$\\
            Compute \method predicted configuration loss, $f_\theta(Y^*, X_s)$ \\
            Update  \method parameters using the regularized network training loss, $\mathcal{L}^{R}_N $ from eq.~\ref{eq:regularized_training_loss} \\
           }  		
   }
   Return $f_\theta$.
	\caption{Training \cmss{ProxyNNs} for Inverse Problems} \label{alg:algorithm_general}
\end{algorithm}

\subsection{Optimization Scheme} 
\label{optimization_scheme}

In general, optimization algorithms based on Newton's methods exhibit rapid convergence but face challenges, such as the computationally expensive calculation of the inverse Hessian matrix and potential numerical instabilities with poorly conditioned Hessians. Quasi-Newton second-order algorithms, like the Broyden–Fletcher–Goldfarb–Shanno algorithm (BFGS), address these issues by approximating the Hessian matrix using information from first-order gradients, requiring less computational cost per iteration. However, they encounter limitations in landscapes with non-linearity, struggling to reach global minima and being susceptible to local minima or saddle points.

To improve convergence in a given non-linear inverse problem, we formulate a two-step optimization strategy utilizing the \cmss{ProxyNNs} trained with regularization hyperparameters $\{ \sigma, \mu\}$ (see section~\ref{optimization_appendix}). In the primary optimization step, we employ \textsc{BFGS} on the configuration loss landscape predicted by the regularized \cmss{ProxyNNs}. The underlying regularization effect is expected to lead to convergence near the global minimum, avoiding regions with local minima and steep gradients in the target landscape. 
To ensure that the predictions can potentially reach arbitrarily high levels of accuracy, a secondary BFGS optimization is introduced. 
We treat the outcome of the primary step convergence as an initial guess for the secondary step, conducted on the ground truth configuration loss function ($\mathcal{L}$). The goal in this step is to attain the optimal solution $X^*$ within the \textit{region of interest}, i.e., the vicinity of the global minimum.

\section{Experimental Setup} \label{sec:experiment}

We perform a series of numerical experiments to investigate (i) the capabilities of \cmss{ProxyNNs} to replicate and regularize the configuration loss function $\mathcal{L}$ and (ii) the optimization performance of regularized \cmss{ProxyNNs} benchmarked against existing iterative methods.

\subsection{Non-Linear Inverse Problems} \label{subsec: problems}
\label{physical_systems}

We evaluate our methodology for the following non-linear physical systems:
\textbf{Burgers\textquotesingle\ equation}: The inviscid Burgers\textquotesingle\ equation \citep{burgers1948mathematical, hopf1950partial} is a fundamental nonlinear first-order hyperbolic partial differential equation that describes the dynamics of a fluid flow and is used in various fields to model a wide range of phenomena \citep{axioms12100982}. The velocity field can exhibit interesting behavior such as the formation of shock waves and turbulence. The equation is written as:
\begin{equation*}
 \frac{\partial u}{\partial t} + u \frac{\partial u}{\partial x} = \nu \frac{\partial^2 u}{\partial x^2}
   \label{eq:burgers_eq}
\end{equation*}
where $u(x,t)$ is the velocity field, and $\nu$ is the kinematic viscosity or diffusivity of the fluid system, which we aim to recover through the solution of the inverse problem.

\textbf{Kuramoto–Sivashinsky equation}: The Kuramoto-Sivashinsky (KS) equation, originally developed to model the unstable behavior of flame fronts \citep{1978PThPS..64..346K}, models a chaotic one-dimensional system $\dot{u}(t) = -\frac{\partial^2 u}{\partial x^2} - \frac{\partial^4 u}{\partial x^4} - u. \nabla u$. We consider a one-parameter inverse problem:
\begin{equation*}
\dot{u}(t) = \alpha \cdot G(x) - \frac{\partial^2 u}{\partial x^2} - \frac{\partial^4 u}{\partial x^4} - \beta \cdot u \cdot \nabla u.
\end{equation*}

where $G(x)$ is a fixed time-independent forcing term; $\alpha, \beta \in \mathbb{R}$ are the unknown parameters governing the evolution. Each inverse problem starts from a randomly generated initial state $u(t = 0)$ and is simulated until $t = 75$, at which point the system becomes chaotic but is still smooth enough to allow for gradient-based optimization. We constrain $\alpha \in [-1, 1]$, $\beta \in [\frac{1}{4}, \frac{3}{4}]$ for numerical stability.

 \begin{figure}[t]
    \centering
    \begin{subfigure}{\textwidth}
        \centering
        \includegraphics[width=0.24\linewidth]{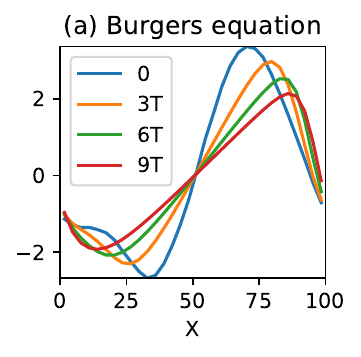}
        \vline
        \includegraphics[width=0.24\linewidth]{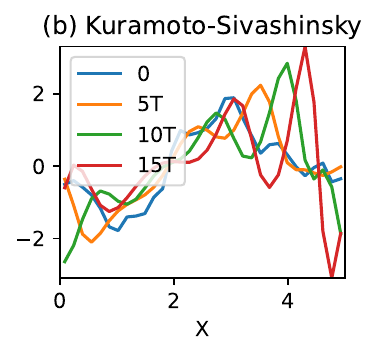}
        \vline
        \includegraphics[width=0.24\linewidth]{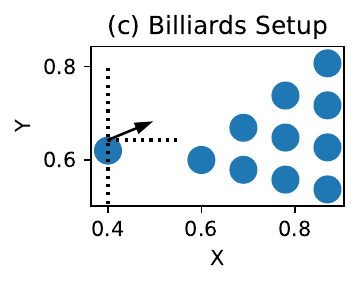}
        \includegraphics[width=0.24\linewidth]{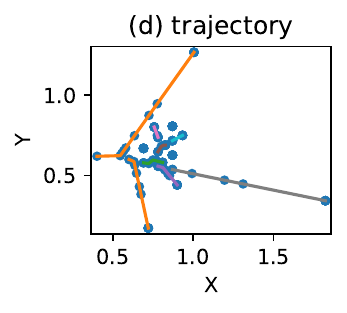}
        
    \end{subfigure}%
    
    \begin{subfigure}{\textwidth}
        \centering
        \includegraphics[width=0.24\linewidth]{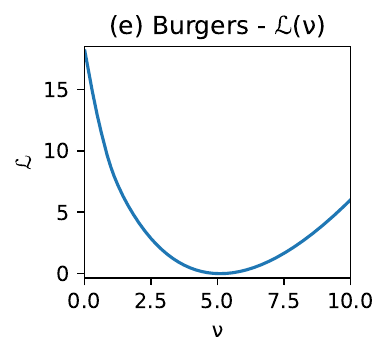}
        \vline
        \includegraphics[width=0.24\linewidth]{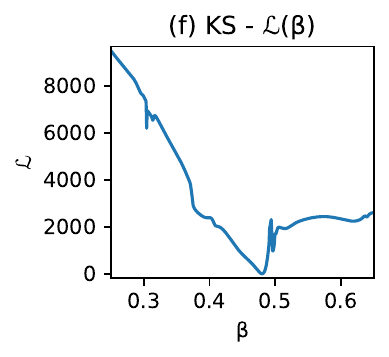}
        \vline
        \includegraphics[width=0.24\linewidth]
        {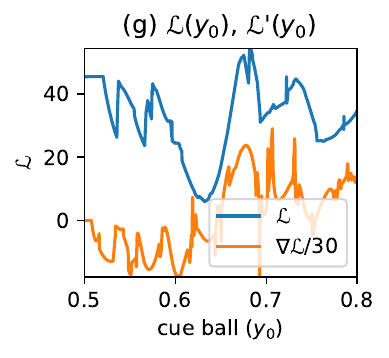}
        \includegraphics[width=0.24\linewidth]{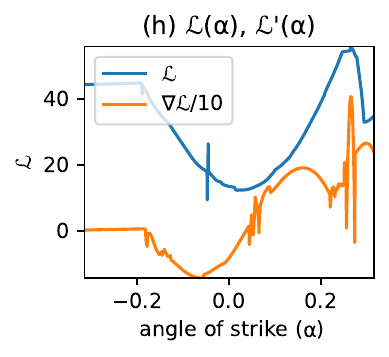}
    \end{subfigure}
    
    \caption{System trajectories and \textit{configuration loss} landscapes ($\mathcal{L}$) for non-linear inverse problems in (a,e) Burgers equation, (b,f) for the Kuramoto-Sivashinsky equation, and (c,d) and (g,h) for the Billiards-2D setup respectively. These landscapes (f-h) encounter convergence issues in iterative optimizers due to local minima and regions of sharp and vanishing gradients.}
    \label{fig:loss_landscapes_all}
\end{figure}

\textbf{Billiards Setup}: Unlike the first two problems which are unidimensional, we also evaluate the performance of the proposed methodology on 2D and 4D classical billiards setup inspired by previous works \citep{hu2020difftaichi}. In the two-dimensional Billiards problem, the angle of strike ($\alpha$) and initial vertical coordinate of the cue ball ($y^0_{cue}$) are treated as target control parameters. We add initial horizontal coordinate ($x^0_{cue}$) and the speed of the cue ball as additional target control parameters for the four-dimensional problem.
The initial location of all the balls (except for the cue ball) is fixed to ensure a fixed initial state $(Y_0)$ for all the possible trajectories. For effective utilization of computational resources,  we extract \textit{key states} and \textit{key times} based on collision occurrences which are subsequently used to generate a linear trajectory.  For a given linear trajectory of states or a subset, the goal is to predict the true angle of strike $(\alpha^*)$ and the initial vertical position of the cue ball $(y^*_{0})$. The temporal evolution of the billiards' trajectory, $Y^*$, encodes the underlying physical laws that govern the collisions and motion of the balls.  

It is important to note that the loss landscapes for the billiards system and the Kuramoto-Sivashinsky equation are very complex (See Figure~\ref{fig:loss_landscapes_all}) with several occurrences of local minima, flat regions, and sharp gradients that are susceptible to slow convergence or convergence to local minima using BFGS.


\subsection{Model Configurations}  \label{subsec: config}


\textbf{Training Data}: We leverage forward simulation datasets for each discussed non-linear physical system. Training the networks with numerically simulated trajectories offers several advantages. Firstly, numerical solutions from the forward solution of PDEs incorporate physical constraints into the system evolution. Secondly, it overcomes the challenges associated with experimental data, which is costly to obtain and often affected by external noise. Lastly, these \cmss{ProxyNNs} can be trained on arbitrarily large amounts of simulation data by solving the underlying PDEs. 

We augment the network inputs $[Y^*, X_s]$ with Fourier features ($\mathcal{F}$) which have been shown to reduce spectral bias in \citet{DBLP:journals/corr/abs-2006-10739}. For example, in each training step in the billiards setup, the network is given access to the full (or a subset) trajectory as one of the inputs $\mathcal{F}(Y^*)$ along with unrelated control parameters $\mathcal{F}(X_s) = \{\alpha_s, \, {y^0_{s}}_{cue}\}$ for 2D and $\mathcal{F}(X_s) = \{{y^0_{s}}_{cue}, {x^0_{s}}_{cue}, \alpha_s, v^0_{cue}\}$ for the 4-D inverse problem to predict the corresponding configuration loss. For 1-D physical systems, we only sample one to two random control parameters for a given starting state (see Table 1) as \cmss{ProxyNNs} are able to generalize even with low-sampling values. Whereas, for Billiards, since there is only one unique starting state, we always sample the target control parameters.

\textbf{Hyperparameters}: We set the learning rate to $0.001$. The batch size, and the number of epochs vary according to the system complexity (see Appendix~\ref{hyperparameter_configuration}). Simulation hyperparameters such as the time step for the Burgers\textquotesingle\ equation and Kuramoto-Shivashinsky equation $(T=0.5)$, coefficient of friction $(\mu = 0.5)$, and elasticity $(e=0.8)$ for the billiards setup were fixed apriori. We trained multiple \cmss{ProxyNNs} with varying hyperparameters $\{\sigma, \mu\}$ to empirically investigate the effect of regularization on the complexity of deep neural networks. \ref{hyperparameter_configuration} summarize the network configurations and all the hyperparameters for the \cmss{ProxyNNs} for different physical systems.
All experiments were conducted on a single Nvidia GTX 1080 Ti using PhiFlow \citep{holl2020phiflow}.\footnote{\url{https://github.com/tum-pbs/PhiFlow}} Training and optimization for each problem takes a maximum of 6-8 hours.

\textbf{Baselines}: We compare our two-step optimization approach using \method with BFGS and Gradient Descent (GD) on the ground truth loss landscapes.

\subsection{Optimization Performance Evaluation} 

For each physical system discussed in (\S~\ref{physical_systems}), we predict the inverse solution, $X_p$, using the optimization scheme discussed in (\S~\ref{optimization_scheme}) on the regularized \cmss{ProxyNNs}. For estimating the convergence accuracy, we evaluate the prediction error ($e$) and the re-simulation error ($r$) as:
\begin{align}
   e &= |X_p - X^*| \\
   r &= ||\mathcal{P}(Y_0, X_p) - \mathcal{P}(Y_0, X^*)||_2^2
   \label{prediction_resim_error}
   \vspace{-0.5cm}
\end{align}
where $X^*$ and $X_p^*$ are the true and predicted values of the control parameters, respectively. The re-simulation error measures the L2 distance between the trajectories originating from the same initial state, $Y_0$ and using the predicted and true values of control parameters.

We further report the convergence accuracy for the top-performing regularized proxy network for all inverse setups discussed in (\S~\ref{physical_systems}) for a range of prediction error thresholds, given by:
\begin{align}
\text{Accuracy} = \frac{\text{Number of predictions with } e \leq \text{threshold}}{\text{Total number of predictions}} \times 100%
\label{accuracy_equation}
\end{align}

\section{Results} \label{sec:results}

In this section, we show that \cmss{ProxyNNs} successfully predict and approximate configuration loss landscapes, $\mathcal{L}$, for each of the physical systems discussed in section~\ref{physical_systems}. Further, we find that the underlying complexity of these networks can be modulated using regularization techniques (\S~\ref{complexity_control}), which subsequently enhances the optimization performance (\S~\ref{optimization_scheme}).

\begin{figure}[t]
    \centering
    \renewcommand\thesubfigure{\alph{subfigure}}
    \begin{subfigure}{0.24\textwidth}
        \includegraphics[width=\linewidth, trim={0 0 0 0.9cm},clip]{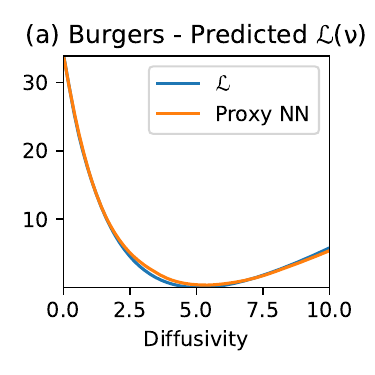}
        \caption{Burgers}
        \end{subfigure}
         \begin{subfigure}{0.24\textwidth}
        \includegraphics[width=\linewidth, trim={0 0 0 0.7cm},clip]{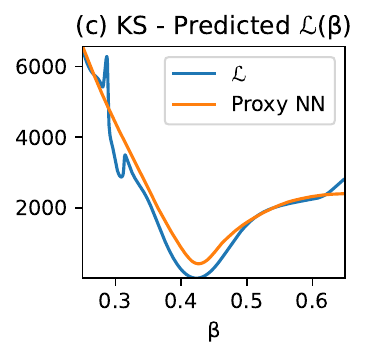} 
         \caption{KS}
        \end{subfigure}
        \vline 
        \begin{subfigure}{0.50\textwidth}
        \includegraphics[width=0.45\linewidth]{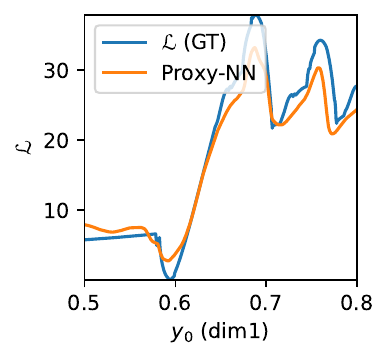} 
         \includegraphics[width=0.45\linewidth]{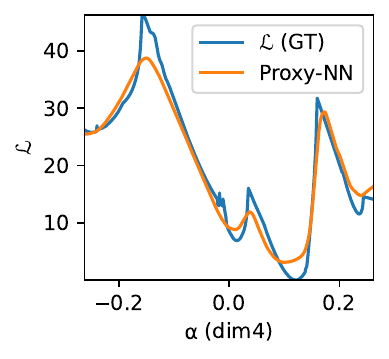}
          \caption{Billiards}
         \end{subfigure}
    \caption{\cmss{ProxyNNs} predict \textit{configuration loss} landscapes $\mathcal{L}$.}
    \label{fig:results_surrogate}
\end{figure}

\paragraph{Proxy Neural Networks Predict Loss Landscapes} 
Figure \ref{fig:results_surrogate} shows that the unregularized \cmss{ProxyNNs} recovers the corresponding \textit{configuration loss} $\mathcal{L}$ for all the inverse problems tested. Although, the ground truth configuration loss function is trivial to optimize in the Burgers equation using \textsc{BFGS}, it acts as a crucial validation that underscores the capabilities of deep networks to predict and approximate the configuration loss landscape governed by non-linear PDEs. 

Unlike the Burgers equation, the configuration loss landscape for the billiards setup and the Kuramoto-Shivashinsky equation is non-trivial with frequent occurrences of local minima, and sharp or vanishing gradients in the target landscape. Figure \ref{fig:results_surrogate}(c) shows the \method predicted landscape for the 4-dimensional billiards setup along $y^0_{cue}$ initial location of the cue ball (true value: $y^*_{cue}=0.59$) and the angle of strike (true value: $\alpha^*=0.134$). We show the other dimensions for the 4D Billiards setup in Appendix Figure~\ref{fig:results_surrogate_4d}. \cmss{ProxyNNs} effectively capture most geometric features of the loss landscape and include essential information about the true parameters at the global minimum. However, addressing the optimization challenges outlined in section~\ref{introduction} remains imperative. This is where the potential to control and regulate the complexity of \cmss{ProxyNNs} presents an opportunity.

\begin{figure}[t]
  \centering
  \begin{subfigure}{\linewidth}
    \includegraphics[width=0.23\linewidth]{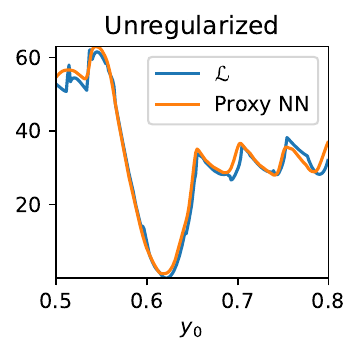} \includegraphics[width=0.23\linewidth]{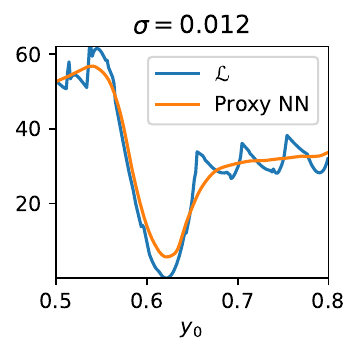}
    \includegraphics[width=0.23\linewidth]{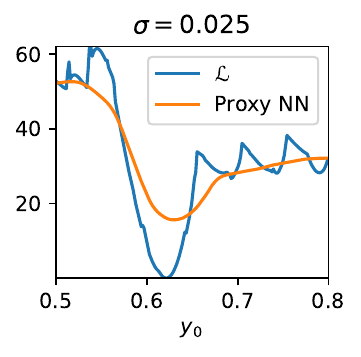}
    \includegraphics[width=0.23\linewidth]{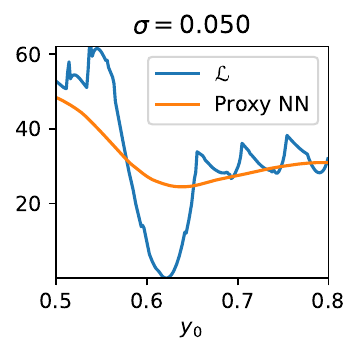}
  \end{subfigure}
  \par
  \begin{subfigure}{\linewidth}
    \includegraphics[width=0.23\linewidth]{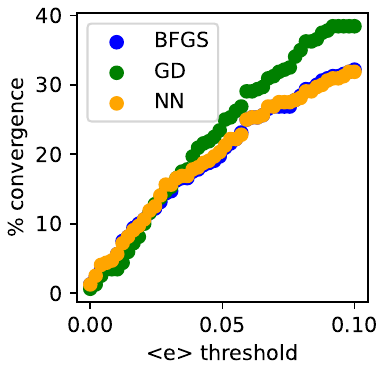}
    \includegraphics[width=0.23\linewidth]{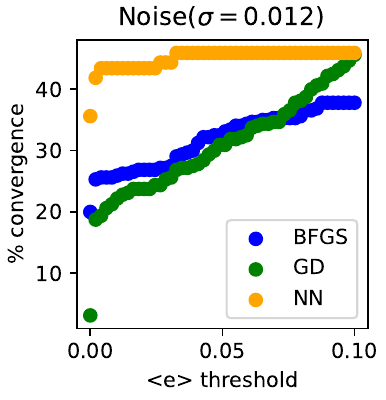}
    \includegraphics[width=0.23\linewidth]{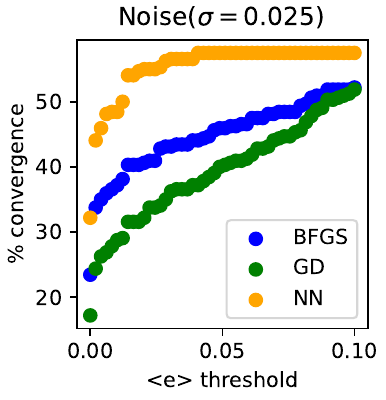}
    \includegraphics[width=0.23\linewidth]{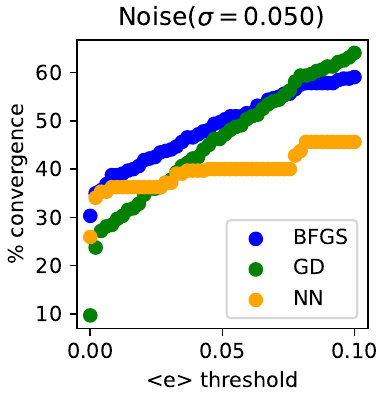}
  \end{subfigure}
  \caption{Predicted Loss Landscapes (top) and Optimization performance (bottom) of \cmss{ProxyNNs} in the Billiards setup trained with various regularization hyperparameter $\sigma=\{0, 0.050\}$. }
  \label{fig:optimization_cases_billiards_all}
\end{figure}

\paragraph{Regularization Reduces Complexity}
In Section~\ref{complexity_control}, we propose to regulate the complexity of the loss landscape predicted by \method using regularization methods. We show the impact of varying the hyperparameter, $\sigma=\{0.0, 0.0125, 0.025, 0.050\}$, on the predicted loss landscape (first row) and convergence accuracy (second row) for the billiards setup (control parameter is the initial position of cue ball $y_{0}$), in Figure~\ref{fig:optimization_cases_billiards_all}. As $\sigma$ increases, the learned loss landscape gets increasingly smoother around the global minimum. Unlike the ``Unregularized'' \method, which yields the same convergence accuracy as \texttt{BFGS}, regularized \cmss{ProxyNNs} consistently improve convergence accuracy over baselines (\texttt{BFGS} and \texttt{GD}). However, for high sigma values $\sigma = 0.05$, the network produces ``too simplified" loss curve, which again yields poor convergence. This demonstrates that the regularization strength plays a key role in simplifying the \method's complexity and can improve convergence accuracy if controlled properly.

 

\begin{figure}[t]
    \centering
    \noindent
    \renewcommand\thesubfigure{\alph{subfigure}}
 \begin{subfigure}{0.24\textwidth}
        \centering
          \includegraphics[width=\linewidth, trim={0 0 0 0.52cm},clip]{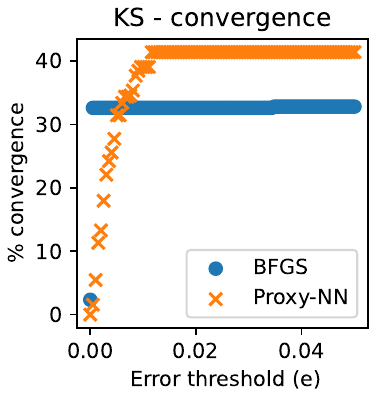}
          \caption{KS}
           \end{subfigure}
    \begin{subfigure}{0.24\textwidth}
        \centering
        \includegraphics[width=\linewidth, trim={0 0 0 0.52cm},clip]{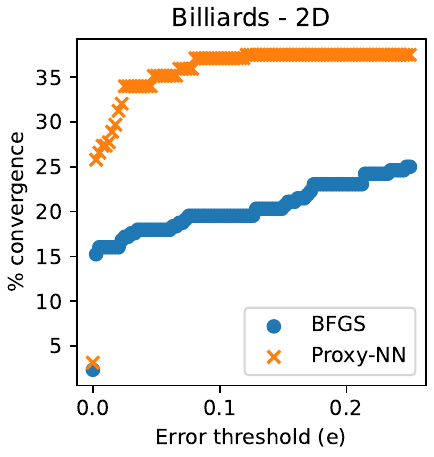}
        \caption{Billiards 2D}
        \end{subfigure}
         \begin{subfigure}{0.24\textwidth}
        \centering
        \includegraphics[width=\linewidth, trim={0 0 0 0.52cm},clip]{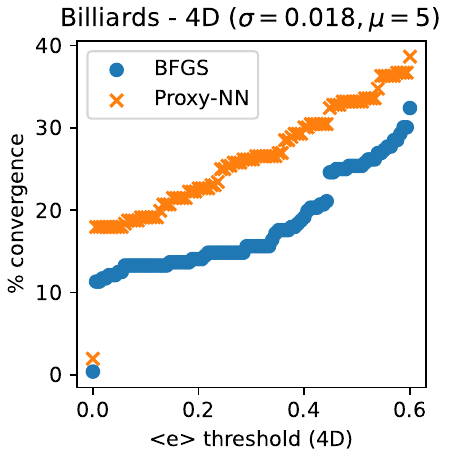}
        \caption{Billiards 4D}
         \end{subfigure}
         \vline
          \begin{subfigure}{0.24\textwidth}
        \centering
        \includegraphics[width=\linewidth, trim={0 0 0 0.52cm},clip]{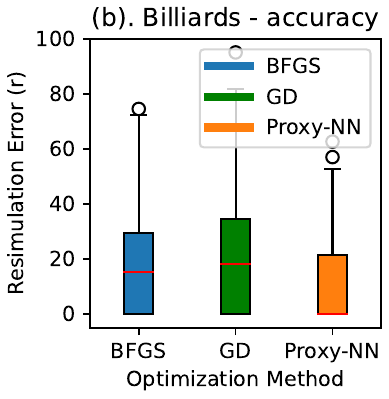}
        \caption{Billiards 2D}
         \end{subfigure}
    \caption{For 256 unique inverse problems for each setup, we report convergence accuracy with varying error thresholds: Regularized \method results in improved accuracy across the board. }
    \label{fig:optimization_results_KS}
\end{figure}

\paragraph{Regularized \cmss{ProxyNNs} Improve Convergence} \label{optimization_results}
\begin{wrapfigure}{0}{0.5\textwidth}
    \includegraphics[width=0.42\linewidth]{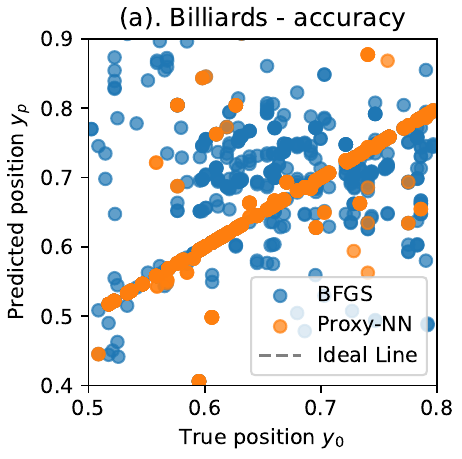}
    \includegraphics[width=0.45\linewidth]{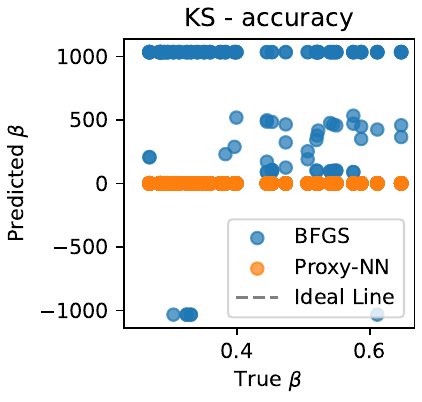}
  \caption{True Vs Predicted Values.}
  \label{fig:truevspredicted}
\end{wrapfigure}
\method consistently yields better convergence across various error thresholds as shown in Figure~\ref{fig:optimization_results_KS}). We also observe that the convergence to the optimal solution almost doubles when using \texttt{BFGS} on \method predicted loss compared to ground truth loss in the 2-D billiards setup. Furthermore, the resimulation error is low for \method compared to other methods. For the Kuramoto–Sivashinsky problem, we can observe that \texttt{BFGS} yields a consistent accuracy score across error thresholds. We find that this is because \texttt{BFGS} either converges on some problems or it does not as shown in Figure~\ref{fig:truevspredicted}, whereas \method results in closer approximation to real $\beta$ values across the board. 

 The improvement in convergence accuracy using \method can be attributed to the smoothening of the loss via the regularization method and the two-step optimization process. The primary optimization step on the regularized \cmss{ProxyNNs} effectively side-steps the local minima, sharp gradients, and flat regions due to underlying regularization. Additionally, the secondary optimization step further enhances the convergence accuracy through the necessary course correction once the vicinity of the global minimum, i.e., the region of interest, is attained.

\section{Related Work}


In recent years, the exploration of machine learning-based methods, including Deep Neural Networks (DNNs) \citep{2020arXiv201104268G,2021arXiv210515044C,2022NJPh...24f3002C, Antil_2023, long2024invertible}, Generative Adversarial Networks (GANs), Variational Autoencoders (VAE), and Bayesian Neural Networks (BNNs), has gained prominence for their potential applications in approximating complex partial differential equations (PDEs) in both forward \citep{PFROMMER2018426, DBLP:journals/corr/abs-1806-02957, MICHOSKI2020193, ANANTHAPADMANABHA2021110194, SHEN2022110460} and inverse problems \citep{anirudh2018unsupervised, sahli2020physics,ray2021solving, zhao2022learning, wu2022inverse,  Jagtap_2022}. Physics-informed neural networks (PINNs) have also emerged as effective tools, integrating established physical laws and constraints into their architecture and training procedures, thereby leveraging prior knowledge of underlying physics to guide the learning process \citep{RAISSI2019686, 10.1115/1.4050542, mao2020physics, math11122723, LI2023111841, 10.1115/1.4053800, fan2020solving, dittmer2020regularization, afkham2021learning, ren2023physicsinformed}. Furthermore, studies have underscored the success of training deep networks using data generated by numerical solvers \citep{10.1145/2939672.2939738, 2020arXiv200408826D, 10.1007/978-3-030-58309-5_20}. 

\citet{ren2021benchmarking} highlighted that by using DNNs as surrogates for the forward model, it is possible to search for good inverse solutions by implementing backpropagation concerning the model inputs. Further, in variations of this approach, termed Neural Adjoint (NA), it has been shown that adding a boundary loss term improves the performance of this method. In the realm of other effective optimization techniques, Bayesian optimization \citep{shahriari2015taking, NIPS2012_05311655,snoek2012practical} is one such promising alternative that provides a robust probabilistic framework for addressing non-linear inverse problems by iteratively exploring and updating the parameter space using observed data. Despite its effectiveness, the method incurs a significant computational cost, primarily attributed to the repeated evaluations of the forward model for each parameter set. Moreover, the efficiency of Bayesian optimization tends to diminish rapidly in high-dimensional inverse loss landscapes. Unlike these approaches, our work aims at directly modeling and using the configuration loss landscapes to identify global minima. By controlling the complexity of the loss landscapes using various regularization methods during training, we identify regions of interest for potential solutions.



\section{Limitations and Future Work}

As emphasized, the influence of regularization on the learning mechanism of the proxy network relies on the inherent non-linearity of the underlying partial differential equation (PDE). This calls for additional exploration to unveil the empirical correlation between the regularization hyperparameters and the geometric complexity of the network. Such an inquiry can be instrumental in fine-tuning and identifying optimal hyperparameters to achieve the desired predictive capabilities. 

Furthermore, while we indeed consider complex 2D and 4D problems, the application of our approach to even more complicated and challenging higher-dimensional problems remains uninvestigated. It would also be interesting to study whether the same underlying network can learn to understand and solve multiple inverse problems together in a multi-task fashion and whether there is any benefit of sharing information across problems. Using neural networks to simulate forward problems is another dimension we do not fully explore as our goal was to focus on modeling the loss landscape induced by the inverse problems.

\section{Conclusion } \label{sec:conclusion}

The framework introduced for generalized configuration loss in this study applies to a range of inverse problems where the system trajectory is observable, and the goal is to identify the inverse parameters that influence the trajectory. 
Neural networks exhibit significant potential in indirectly predicting intricate loss relations between system observables and randomly sampled target parameters. Additionally, \cmss{ProxyNNs} trained with Fourier feature mappings, showcase effective utilization of model capacity, faster convergence, and enhanced sensitivity to the regularization methods.

Additionally, we explore how the learning dynamics of \cmss{ProxyNNs} can be shaped using regularization methods, introducing a regularization pressure aimed at managing the inherent complexity. This ability to control complexity positions solutions based on \cmss{ProxyNNs} as effective choices for addressing inverse problems. We observe that the regularization methods discussed in section~\ref{complexity_control} consistently simplify the geometrically intricate features of the underlying loss landscape, allowing for fine-tuning through corresponding hyperparameters. 

The observed optimization performance of BFGS on the regularized \cmss{ProxyNNs} in section~\ref{optimization_results} provides empirical support that the learning dynamics of neural networks can be harnessed to exert control over their inherent complexity. This, in turn, proves beneficial for the optimization process within existing iterative optimizers, enabling them to predict optimal parameters more effectively.



\nocite{langley00}
\bibliography{main.bib}

\begin{thebibliography}{56}
\providecommand{\natexlab}[1]{#1}
\providecommand{\url}[1]{\texttt{#1}}
\expandafter\ifx\csname urlstyle\endcsname\relax
  \providecommand{\doi}[1]{doi: #1}\else
  \providecommand{\doi}{doi: \begingroup \urlstyle{rm}\Url}\fi

\bibitem[Afkham et~al.(2021)Afkham, Chung, and Chung]{afkham2021learning}
Afkham, B.~M., Chung, J., and Chung, M.
\newblock Learning regularization parameters of inverse problems via deep neural networks.
\newblock \emph{Inverse Problems}, 37\penalty0 (10):\penalty0 105017, 2021.

\bibitem[{Anantha Padmanabha} \& Zabaras(2021){Anantha Padmanabha} and Zabaras]{ANANTHAPADMANABHA2021110194}
{Anantha Padmanabha}, G. and Zabaras, N.
\newblock Solving inverse problems using conditional invertible neural networks.
\newblock \emph{Journal of Computational Physics}, 433:\penalty0 110194, 2021.
\newblock ISSN 0021-9991.
\newblock \doi{https://doi.org/10.1016/j.jcp.2021.110194}.
\newblock URL \url{https://www.sciencedirect.com/science/article/pii/S0021999121000899}.

\bibitem[Anirudh et~al.(2018)Anirudh, Thiagarajan, Kailkhura, and Bremer]{anirudh2018unsupervised}
Anirudh, R., Thiagarajan, J.~J., Kailkhura, B., and Bremer, T.
\newblock An unsupervised approach to solving inverse problems using generative adversarial networks.
\newblock \emph{arXiv preprint arXiv:1805.07281}, 2018.

\bibitem[Antil et~al.(2023)Antil, Elman, Onwunta, and Verma]{Antil_2023}
Antil, H., Elman, H.~C., Onwunta, A., and Verma, D.
\newblock A deep neural network approach for parameterized pdes and bayesian inverse problems.
\newblock \emph{Machine Learning: Science and Technology}, 4\penalty0 (3):\penalty0 035015, aug 2023.
\newblock \doi{10.1088/2632-2153/ace67c}.
\newblock URL \url{https://dx.doi.org/10.1088/2632-2153/ace67c}.

\bibitem[Blanc et~al.(2020)Blanc, Gupta, Valiant, and Valiant]{blanc2020implicit}
Blanc, G., Gupta, N., Valiant, G., and Valiant, P.
\newblock Implicit regularization for deep neural networks driven by an ornstein-uhlenbeck like process, 2020.

\bibitem[Broyden et~al.(1970)Broyden, Fletcher, Goldfarb, and Shanno]{broyden1970convergence}
Broyden, C., Fletcher, R., Goldfarb, D., and Shanno, D.
\newblock Convergence properties of a class of quasi-newton methods in optimization.
\newblock \emph{Journal of Mathematical Programming}, 6\penalty0 (2):\penalty0 163--175, 1970.

\bibitem[Buhrmester et~al.(2021)Buhrmester, Münch, and Arens]{make3040048}
Buhrmester, V., Münch, D., and Arens, M.
\newblock Analysis of explainers of black box deep neural networks for computer vision: A survey.
\newblock \emph{Machine Learning and Knowledge Extraction}, 3\penalty0 (4):\penalty0 966--989, 2021.
\newblock ISSN 2504-4990.
\newblock \doi{10.3390/make3040048}.
\newblock URL \url{https://www.mdpi.com/2504-4990/3/4/48}.

\bibitem[Burgers(1948)]{burgers1948mathematical}
Burgers, J.~M.
\newblock A mathematical model illustrating the theory of turbulence.
\newblock \emph{Advances in applied mechanics}, 1:\penalty0 171--199, 1948.

\bibitem[Cai et~al.(2021)Cai, Wang, Wang, Perdikaris, and Karniadakis]{10.1115/1.4050542}
Cai, S., Wang, Z., Wang, S., Perdikaris, P., and Karniadakis, G.~E.
\newblock {Physics-Informed Neural Networks for Heat Transfer Problems}.
\newblock \emph{Journal of Heat Transfer}, 143\penalty0 (6), 04 2021.
\newblock ISSN 0022-1481.
\newblock \doi{10.1115/1.4050542}.
\newblock URL \url{https://doi.org/10.1115/1.4050542}.
\newblock 060801.

\bibitem[{Cao} et~al.(2022){Cao}, {Xie}, {Zhang}, {Hou}, {Zhang}, and {Zeng}]{2022NJPh...24f3002C}
{Cao}, N., {Xie}, J., {Zhang}, A., {Hou}, S.-Y., {Zhang}, L., and {Zeng}, B.
\newblock {Neural networks for quantum inverse problems}.
\newblock \emph{New Journal of Physics}, 24\penalty0 (6):\penalty0 063002, June 2022.
\newblock \doi{10.1088/1367-2630/ac706c}.

\bibitem[{Chouzenoux} et~al.(2021){Chouzenoux}, {Della Valle}, and {Pesquet}]{2021arXiv210515044C}
{Chouzenoux}, E., {Della Valle}, C., and {Pesquet}, J.-C.
\newblock {Inversion of Integral Models: a Neural Network Approach}.
\newblock \emph{arXiv e-prints}, art. arXiv:2105.15044, May 2021.
\newblock \doi{10.48550/arXiv.2105.15044}.

\bibitem[Dherin et~al.(2022)Dherin, Munn, Rosca, and Barrett]{dherin2022neural}
Dherin, B., Munn, M., Rosca, M., and Barrett, D. G.~T.
\newblock Why neural networks find simple solutions: the many regularizers of geometric complexity, 2022.

\bibitem[{Dias Ribeiro} et~al.(2020){Dias Ribeiro}, {Rehman}, {Ahmed}, and {Dengel}]{2020arXiv200408826D}
{Dias Ribeiro}, M., {Rehman}, A., {Ahmed}, S., and {Dengel}, A.
\newblock {DeepCFD: Efficient Steady-State Laminar Flow Approximation with Deep Convolutional Neural Networks}.
\newblock \emph{arXiv e-prints}, art. arXiv:2004.08826, April 2020.
\newblock \doi{10.48550/arXiv.2004.08826}.

\bibitem[Dittmer et~al.(2020)Dittmer, Kluth, Maass, and Otero~Baguer]{dittmer2020regularization}
Dittmer, S., Kluth, T., Maass, P., and Otero~Baguer, D.
\newblock Regularization by architecture: A deep prior approach for inverse problems.
\newblock \emph{Journal of Mathematical Imaging and Vision}, 62:\penalty0 456--470, 2020.

\bibitem[Fan et~al.(2020)Fan, Xu, Pathak, and Darve]{fan2020solving}
Fan, T., Xu, K., Pathak, J., and Darve, E.
\newblock Solving inverse problems in steady-state navier-stokes equations using deep neural networks, 2020.

\bibitem[Fung et~al.(2021)Fung, Zhang, Hu, Ganesh, and Sumpter]{Fung2021}
Fung, V., Zhang, J., Hu, G., Ganesh, P., and Sumpter, B.~G.
\newblock Inverse design of two-dimensional materials with invertible neural networks.
\newblock \emph{npj Computational Materials}, 7\penalty0 (1):\penalty0 200, Dec 2021.
\newblock ISSN 2057-3960.
\newblock \doi{10.1038/s41524-021-00670-x}.
\newblock URL \url{https://doi.org/10.1038/s41524-021-00670-x}.

\bibitem[{Genzel} et~al.(2020){Genzel}, {Macdonald}, and {M{\"a}rz}]{2020arXiv201104268G}
{Genzel}, M., {Macdonald}, J., and {M{\"a}rz}, M.
\newblock {Solving Inverse Problems With Deep Neural Networks -- Robustness Included?}
\newblock \emph{arXiv e-prints}, art. arXiv:2011.04268, November 2020.
\newblock \doi{10.48550/arXiv.2011.04268}.

\bibitem[Goodfellow et~al.(2016)Goodfellow, Bengio, and Courville]{Goodfellow-et-al-2016}
Goodfellow, I., Bengio, Y., and Courville, A.
\newblock \emph{Deep Learning}.
\newblock MIT Press, 2016.
\newblock \url{http://www.deeplearningbook.org}.

\bibitem[Gramacy \& Lee(2010)Gramacy and Lee]{gramacy2010optimization}
Gramacy, R.~B. and Lee, H. K.~H.
\newblock Optimization under unknown constraints, 2010.

\bibitem[Guo et~al.(2016)Guo, Li, and Iorio]{10.1145/2939672.2939738}
Guo, X., Li, W., and Iorio, F.
\newblock Convolutional neural networks for steady flow approximation.
\newblock In \emph{Proceedings of the 22nd ACM SIGKDD International Conference on Knowledge Discovery and Data Mining}, KDD '16, pp.\  481–490, New York, NY, USA, 2016. Association for Computing Machinery.
\newblock ISBN 9781450342322.
\newblock \doi{10.1145/2939672.2939738}.
\newblock URL \url{https://doi.org/10.1145/2939672.2939738}.

\bibitem[Holl et~al.(2020)Holl, Koltun, Um, and Thuerey]{holl2020phiflow}
Holl, P., Koltun, V., Um, K., and Thuerey, N.
\newblock phiflow: A differentiable pde solving framework for deep learning via physical simulations.
\newblock In \emph{NeurIPS workshop}, volume~2, 2020.

\bibitem[Hopf(1950)]{hopf1950partial}
Hopf, E.
\newblock The partial differential equation.
\newblock 1950.

\bibitem[Hu et~al.(2020)Hu, Anderson, Li, Sun, Carr, Ragan-Kelley, and Durand]{hu2020difftaichi}
Hu, Y., Anderson, L., Li, T.-M., Sun, Q., Carr, N., Ragan-Kelley, J., and Durand, F.
\newblock Difftaichi: Differentiable programming for physical simulation, 2020.

\bibitem[Huang et~al.(2005)Huang, Xiang, Du, and Cao]{Sixun_Huang_2005}
Huang, S., Xiang, J., Du, H., and Cao, X.
\newblock Inverse problems in atmospheric science and their application.
\newblock \emph{Journal of Physics: Conference Series}, 12\penalty0 (1):\penalty0 45, jan 2005.
\newblock \doi{10.1088/1742-6596/12/1/005}.
\newblock URL \url{https://dx.doi.org/10.1088/1742-6596/12/1/005}.

\bibitem[Jagtap et~al.(2022)Jagtap, Mao, Adams, and Karniadakis]{Jagtap_2022}
Jagtap, A.~D., Mao, Z., Adams, N., and Karniadakis, G.~E.
\newblock Physics-informed neural networks for inverse problems in supersonic flows.
\newblock \emph{Journal of Computational Physics}, 466:\penalty0 111402, oct 2022.
\newblock \doi{10.1016/j.jcp.2022.111402}.
\newblock URL \url{https://doi.org/10.1016%2Fj.jcp.2022.111402}.

\bibitem[Kingma \& Ba(2017)Kingma and Ba]{kingma2017adam}
Kingma, D.~P. and Ba, J.
\newblock Adam: A method for stochastic optimization, 2017.

\bibitem[{Kuramoto}(1978)]{1978PThPS..64..346K}
{Kuramoto}, Y.
\newblock {Diffusion-Induced Chaos in Reaction Systems}.
\newblock \emph{Progress of Theoretical Physics Supplement}, 64:\penalty0 346--367, January 1978.
\newblock \doi{10.1143/PTPS.64.346}.

\bibitem[Langley(2000)]{langley00}
Langley, P.
\newblock Crafting papers on machine learning.
\newblock In Langley, P. (ed.), \emph{Proceedings of the 17th International Conference on Machine Learning (ICML 2000)}, pp.\  1207--1216, Stanford, CA, 2000. Morgan Kaufmann.

\bibitem[Li et~al.(2023)Li, Wang, and Yan]{LI2023111841}
Li, Y., Wang, Y., and Yan, L.
\newblock Surrogate modeling for bayesian inverse problems based on physics-informed neural networks.
\newblock \emph{Journal of Computational Physics}, 475:\penalty0 111841, 2023.
\newblock ISSN 0021-9991.
\newblock \doi{https://doi.org/10.1016/j.jcp.2022.111841}.
\newblock URL \url{https://www.sciencedirect.com/science/article/pii/S0021999122009044}.

\bibitem[Long \& Zhe(2024)Long and Zhe]{long2024invertible}
Long, D. and Zhe, S.
\newblock Invertible fourier neural operators for tackling both forward and inverse problems, 2024.

\bibitem[Mao et~al.(2020)Mao, Jagtap, and Karniadakis]{mao2020physics}
Mao, Z., Jagtap, A.~D., and Karniadakis, G.~E.
\newblock Physics-informed neural networks for high-speed flows.
\newblock \emph{Computer Methods in Applied Mechanics and Engineering}, 360:\penalty0 112789, 2020.

\bibitem[McCann et~al.(2017)McCann, Jin, and Unser]{8103129}
McCann, M.~T., Jin, K.~H., and Unser, M.
\newblock Convolutional neural networks for inverse problems in imaging: A review.
\newblock \emph{IEEE Signal Processing Magazine}, 34\penalty0 (6):\penalty0 85--95, 2017.
\newblock \doi{10.1109/MSP.2017.2739299}.

\bibitem[Michoski et~al.(2020)Michoski, Milosavljević, Oliver, and Hatch]{MICHOSKI2020193}
Michoski, C., Milosavljević, M., Oliver, T., and Hatch, D.~R.
\newblock Solving differential equations using deep neural networks.
\newblock \emph{Neurocomputing}, 399:\penalty0 193--212, 2020.
\newblock ISSN 0925-2312.
\newblock \doi{https://doi.org/10.1016/j.neucom.2020.02.015}.
\newblock URL \url{https://www.sciencedirect.com/science/article/pii/S0925231220301909}.

\bibitem[Mohammad-Djafari(2021)]{mohammad2021regularization}
Mohammad-Djafari, A.
\newblock Regularization, bayesian inference, and machine learning methods for inverse problems.
\newblock \emph{Entropy}, 23\penalty0 (12):\penalty0 1673, 2021.

\bibitem[Nabian \& Meidani(2018)Nabian and Meidani]{DBLP:journals/corr/abs-1806-02957}
Nabian, M.~A. and Meidani, H.
\newblock A deep neural network surrogate for high-dimensional random partial differential equations.
\newblock \emph{CoRR}, abs/1806.02957, 2018.
\newblock URL \url{http://arxiv.org/abs/1806.02957}.

\bibitem[Oommen \& Srinivasan(2022)Oommen and Srinivasan]{10.1115/1.4053800}
Oommen, V. and Srinivasan, B.
\newblock {Solving Inverse Heat Transfer Problems Without Surrogate Models: A Fast, Data-Sparse, Physics Informed Neural Network Approach}.
\newblock \emph{Journal of Computing and Information Science in Engineering}, 22\penalty0 (4):\penalty0 041012, 03 2022.
\newblock ISSN 1530-9827.
\newblock \doi{10.1115/1.4053800}.
\newblock URL \url{https://doi.org/10.1115/1.4053800}.

\bibitem[Oulghelou et~al.(2022)Oulghelou, Beghein, and Allery]{OULGHELOU2022105490}
Oulghelou, M., Beghein, C., and Allery, C.
\newblock A surrogate optimization approach for inverse problems: Application to turbulent mixed-convection flows.
\newblock \emph{Computers \& Fluids}, 241:\penalty0 105490, 2022.
\newblock ISSN 0045-7930.
\newblock \doi{https://doi.org/10.1016/j.compfluid.2022.105490}.
\newblock URL \url{https://www.sciencedirect.com/science/article/pii/S0045793022001311}.

\bibitem[Pfrommer et~al.(2018)Pfrommer, Zimmerling, Liu, Kärger, Henning, and Beyerer]{PFROMMER2018426}
Pfrommer, J., Zimmerling, C., Liu, J., Kärger, L., Henning, F., and Beyerer, J.
\newblock Optimisation of manufacturing process parameters using deep neural networks as surrogate models.
\newblock \emph{Procedia CIRP}, 72:\penalty0 426--431, 2018.
\newblock ISSN 2212-8271.
\newblock \doi{https://doi.org/10.1016/j.procir.2018.03.046}.
\newblock URL \url{https://www.sciencedirect.com/science/article/pii/S221282711830146X}.
\newblock 51st CIRP Conference on Manufacturing Systems.

\bibitem[Raissi et~al.(2019)Raissi, Perdikaris, and Karniadakis]{RAISSI2019686}
Raissi, M., Perdikaris, P., and Karniadakis, G.
\newblock Physics-informed neural networks: A deep learning framework for solving forward and inverse problems involving nonlinear partial differential equations.
\newblock \emph{Journal of Computational Physics}, 378:\penalty0 686--707, 2019.
\newblock ISSN 0021-9991.
\newblock \doi{https://doi.org/10.1016/j.jcp.2018.10.045}.
\newblock URL \url{https://www.sciencedirect.com/science/article/pii/S0021999118307125}.

\bibitem[Ray(2021)]{ray2021solving}
Ray, D.
\newblock Solving physics-based inverse problems using gans.
\newblock 2021.

\bibitem[Ren et~al.(2023)Ren, Rao, Sun, and Liu]{ren2023physicsinformed}
Ren, P., Rao, C., Sun, H., and Liu, Y.
\newblock Physics-informed neural network for seismic wave inversion in layered semi-infinite domain, 2023.

\bibitem[Ren et~al.(2021)Ren, Padilla, and Malof]{ren2021benchmarking}
Ren, S., Padilla, W., and Malof, J.
\newblock Benchmarking deep inverse models over time, and the neural-adjoint method, 2021.

\bibitem[Ruder(2017)]{ruder2017overview}
Ruder, S.
\newblock An overview of gradient descent optimization algorithms, 2017.

\bibitem[Sahli~Costabal et~al.(2020)Sahli~Costabal, Yang, Perdikaris, Hurtado, and Kuhl]{sahli2020physics}
Sahli~Costabal, F., Yang, Y., Perdikaris, P., Hurtado, D.~E., and Kuhl, E.
\newblock Physics-informed neural networks for cardiac activation mapping.
\newblock \emph{Frontiers in Physics}, 8:\penalty0 42, 2020.

\bibitem[Savović et~al.(2023)Savović, Ivanović, and Min]{axioms12100982}
Savović, S., Ivanović, M., and Min, R.
\newblock A comparative study of the explicit finite difference method and physics-informed neural networks for solving the burgers; equation.
\newblock \emph{Axioms}, 12\penalty0 (10), 2023.
\newblock ISSN 2075-1680.
\newblock \doi{10.3390/axioms12100982}.
\newblock URL \url{https://www.mdpi.com/2075-1680/12/10/982}.

\bibitem[Shahriari et~al.(2015)Shahriari, Swersky, Wang, Adams, and De~Freitas]{shahriari2015taking}
Shahriari, B., Swersky, K., Wang, Z., Adams, R.~P., and De~Freitas, N.
\newblock Taking the human out of the loop: A review of bayesian optimization.
\newblock \emph{Proceedings of the IEEE}, 104\penalty0 (1):\penalty0 148--175, 2015.

\bibitem[Shen et~al.(2022)Shen, Li, Zha, Li, and Liu]{SHEN2022110460}
Shen, L., Li, D., Zha, W., Li, X., and Liu, X.
\newblock Surrogate modeling for porous flow using deep neural networks.
\newblock \emph{Journal of Petroleum Science and Engineering}, 213:\penalty0 110460, 2022.
\newblock ISSN 0920-4105.
\newblock \doi{https://doi.org/10.1016/j.petrol.2022.110460}.
\newblock URL \url{https://www.sciencedirect.com/science/article/pii/S092041052200345X}.

\bibitem[Snoek et~al.(2012{\natexlab{a}})Snoek, Larochelle, and Adams]{NIPS2012_05311655}
Snoek, J., Larochelle, H., and Adams, R.~P.
\newblock Practical bayesian optimization of machine learning algorithms.
\newblock In Pereira, F., Burges, C., Bottou, L., and Weinberger, K. (eds.), \emph{Advances in Neural Information Processing Systems}, volume~25. Curran Associates, Inc., 2012{\natexlab{a}}.
\newblock URL \url{https://proceedings.neurips.cc/paper_files/paper/2012/file/05311655a15b75fab86956663e1819cd-Paper.pdf}.

\bibitem[Snoek et~al.(2012{\natexlab{b}})Snoek, Larochelle, and Adams]{snoek2012practical}
Snoek, J., Larochelle, H., and Adams, R.~P.
\newblock Practical bayesian optimization of machine learning algorithms, 2012{\natexlab{b}}.

\bibitem[Tancik et~al.(2020)Tancik, Srinivasan, Mildenhall, Fridovich{-}Keil, Raghavan, Singhal, Ramamoorthi, Barron, and Ng]{DBLP:journals/corr/abs-2006-10739}
Tancik, M., Srinivasan, P.~P., Mildenhall, B., Fridovich{-}Keil, S., Raghavan, N., Singhal, U., Ramamoorthi, R., Barron, J.~T., and Ng, R.
\newblock Fourier features let networks learn high-frequency functions in low dimensional domains.
\newblock \emph{CoRR}, abs/2006.10739, 2020.
\newblock URL \url{https://arxiv.org/abs/2006.10739}.

\bibitem[Winovich et~al.(2019)Winovich, Ramani, and Lin]{WINOVICH2019263}
Winovich, N., Ramani, K., and Lin, G.
\newblock Convpde-uq: Convolutional neural networks with quantified uncertainty for heterogeneous elliptic partial differential equations on varied domains.
\newblock \emph{Journal of Computational Physics}, 394:\penalty0 263--279, 2019.
\newblock ISSN 0021-9991.
\newblock \doi{https://doi.org/10.1016/j.jcp.2019.05.026}.
\newblock URL \url{https://www.sciencedirect.com/science/article/pii/S0021999119303572}.

\bibitem[Wolf et~al.(2020)Wolf, Donzallaz, Jost, Hayoz, Commend, Hennebert, and Kuonen]{10.1007/978-3-030-58309-5_20}
Wolf, B., Donzallaz, J., Jost, C., Hayoz, A., Commend, S., Hennebert, J., and Kuonen, P.
\newblock Using cnns to optimize numerical simulations in geotechnical engineering.
\newblock In \emph{Artificial Neural Networks in Pattern Recognition: 9th IAPR TC3 Workshop, ANNPR 2020, Winterthur, Switzerland, September 2–4, 2020, Proceedings}, pp.\  247–256, Berlin, Heidelberg, 2020. Springer-Verlag.
\newblock ISBN 978-3-030-58308-8.
\newblock \doi{10.1007/978-3-030-58309-5_20}.
\newblock URL \url{https://doi.org/10.1007/978-3-030-58309-5_20}.

\bibitem[Wu et~al.(2022)Wu, O'Malley, Golden, and Vesselinov]{wu2022inverse}
Wu, H., O'Malley, D., Golden, J.~K., and Vesselinov, V.~V.
\newblock Inverse analysis with variational autoencoders: A comparison of shallow and deep networks.
\newblock \emph{Journal of Machine Learning for Modeling and Computing}, 3\penalty0 (2), 2022.

\bibitem[Würth et~al.(2023)Würth, Krauß, Zimmerling, and Kärger]{WURTH2023112034}
Würth, T., Krauß, C., Zimmerling, C., and Kärger, L.
\newblock Physics-informed neural networks for data-free surrogate modelling and engineering optimization – an example from composite manufacturing.
\newblock \emph{Materials \& Design}, 231:\penalty0 112034, 2023.
\newblock ISSN 0264-1275.
\newblock \doi{https://doi.org/10.1016/j.matdes.2023.112034}.
\newblock URL \url{https://www.sciencedirect.com/science/article/pii/S0264127523004495}.

\bibitem[Zhao et~al.(2022)Zhao, Lindell, and Wetzstein]{zhao2022learning}
Zhao, Q., Lindell, D.~B., and Wetzstein, G.
\newblock Learning to solve pde-constrained inverse problems with graph networks, 2022.

\bibitem[Zhi et~al.(2023)Zhi, Wu, Qi, Zhu, Wu, and Wu]{math11122723}
Zhi, P., Wu, Y., Qi, C., Zhu, T., Wu, X., and Wu, H.
\newblock Surrogate-based physics-informed neural networks for elliptic partial differential equations.
\newblock \emph{Mathematics}, 11\penalty0 (12), 2023.
\newblock ISSN 2227-7390.
\newblock \doi{10.3390/math11122723}.
\newblock URL \url{https://www.mdpi.com/2227-7390/11/12/2723}.

\end{thebibliography}
\bibliographystyle{icml2021}

\appendix






\section{Hyperparameter Configuration} \label{sec:hyper}
\label{hyperparameter_configuration}

The network hyperparameters for each of the settings and the dataset details are summarized below:

\begin{table}[h]
\small
\centering
\setlength{\tabcolsep}{2pt}
\scalebox{0.70}{
\begin{tabular}{cccccccc}
\toprule
\multirow{2}{*}{System} & \multirow{2}{*}{\# Input Dim} & \multirow{2}{*}{Architecture} & \multirow{2}{*}{\# Parameters} & \multirow{2}{*}{Dataset Size} & \multirow{2}{*}{Batch Size} & \multirow{2}{*}{Sampling} &   \multirow{2}{*}{($\sigma$, $\mu$)} \\
\addlinespace[0.2cm]
& & & ($\times 10^3$) & ($\times 10^3$) & &  Rate & \\
\midrule
\addlinespace[0.1cm]
Rastrigin  & 1 & DenseNet & 49.7 & NR & 256 & NA & ([0.20, 0.50], 1)\\
Test function & & [64,128,128,128,64] &  \\
\addlinespace[0.3cm]
Gramacy \& Lee  & 1 & DenseNet & 82.7 & NR & 256 & NA & ([0.17, 0.27], 20) \\
Test function & & [64,128,256,128,64] &  \\
\addlinespace[0.3cm]
Burgers\textquotesingle\  & 2 & ConvNet & 107.6 & 32 & 128 & 2 & NR  \\
equation & & [128,128,128] & & \\
\addlinespace[0.3cm]
Kuramoto-Sivashinsky & 16 & ConvNet with $\mathcal{F}$ & 779.5 & 1280 & 256 & 2 & (\{0.005, 0.021, 0.1\}, 5) \\
equation& & [32, 64, 128, 256, 128, 64,32] & & \\
\addlinespace[0.3cm]
Billiards-2D  & 40 & ConvNet with $\mathcal{F} $& 740.9 & 640 & 128 & 640$k$ & (\{0.013, 0.018, 0.025, 0.027, 0.050\}, 5)\\
Setup & & [32,64,128,256,128,64,32] & & \\
\addlinespace[0.3cm]
Billiards-4D  & 64 & ConvNet with $\mathcal{F} $& 740.9 & 2560 & 256 & 2560$k$ & (\{0.013, 0.018, 0.025, 0.027, 0.050\}, 5)\\
Setup & & [32,64,128,256,128,64,32] & & \\
\bottomrule
\end{tabular}}
\caption{Training details for test functions and physical systems. NR = Not Required. In both 2D and 4D Billiards, there is only one unique initial state.}
\label{tab:surrogate_nn_details}
\end{table}

\section{Optimization policy and results}
\label{optimization_appendix}

\begin{figure}[h]
    \centering
    \includegraphics[width=0.50\textwidth]{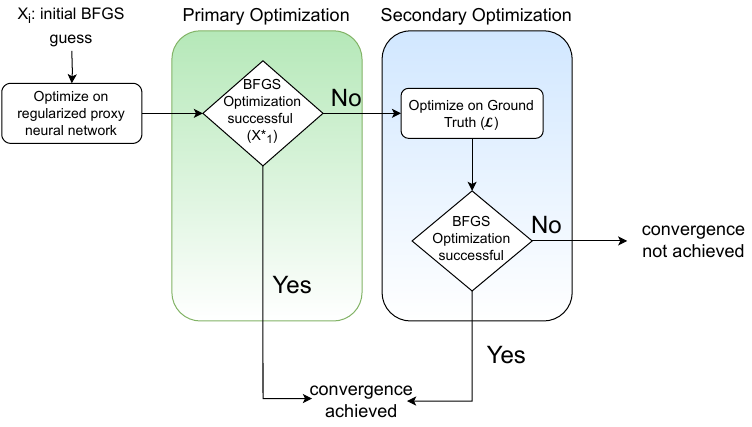}
    \caption{Two-Step optimization policy for \cmss{ProxyNNs}. The primary step aims to converge in the vicinity of the global minimum which is treated as initial guess for the secondary optimization step.}
    \label{fig:optimization_scheme}
\end{figure}

The two-step optimization strategy and performance are depicted in figure~\ref{fig:optimization_scheme}. We aim to locate the \textit{``region of interest"} in the control parameter search space $Z$ that contains the global minimum through the primary optimization step. This effectively narrows down the search limit to the vicinity of the global minimum. Furthermore, starting from the primary convergence point ($X_1^*$) we achieve convergence close to the global minimum ($X_p^*$) through final optimization on the ground truth \textit{configuration loss}, $\mathcal{L}$.\\

\section{\method predicted \textit{Configuration Loss} landscapes for 4D Billiards}

\begin{figure}[h]
    \centering
    \begin{subfigure}{\textwidth}
        \centering
      \includegraphics[width=0.23\textwidth]{figures/billiards_4d_dim1_pos_y_gt_s_003_.pdf} 
        \label{fig:3d_billiards_predicted_loss_1d}
        \includegraphics[width=0.23\textwidth]{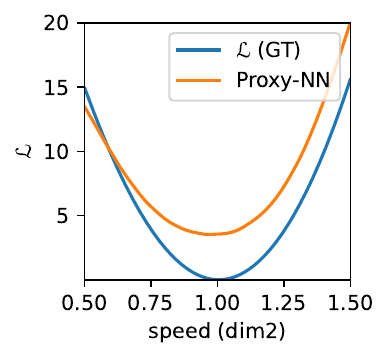}
        \label{fig:3a_burgers_proxy_NN}
        \includegraphics[width=0.23\textwidth]{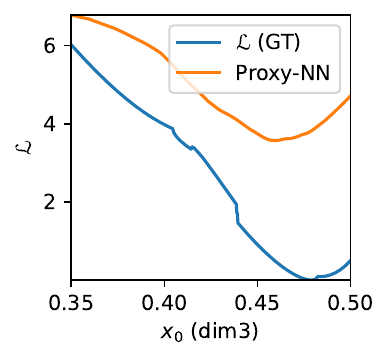} 
        \label{fig:3c_KS_predicted_loss}
         \includegraphics[width=0.23\textwidth]{figures/billiards_4d_dim4_alpha_gt_s_002_.pdf} 
        \label{fig:3b_billiards_predicted_loss_2d}
    \end{subfigure}
    \caption{Proxy networks predicts \textit{configuration loss} landscape $\mathcal{L}$ in the 4-D Billiards Inverse Setup.}
    \label{fig:results_surrogate_4d}
\end{figure}

\end{document}